\documentclass[a4paper]{svproc}
\usepackage{url}

\usepackage[frozencache]{minted}
\usepackage{amsmath}
\usepackage{amsfonts}
\usepackage{amssymb}
\usepackage{graphicx}
\usepackage{tabularx}
\usepackage{caption}
\usepackage{algpseudocode}
\usepackage{enumitem}
\usepackage{wrapfig}
\usepackage{verbatim}
\usepackage{color}
\usepackage{xcolor}
\usepackage{textcomp}
\usepackage{subfig}

\graphicspath{ {fig/} }

\newcolumntype{L}[1]{>{\raggedright\arraybackslash}p{#1}}
\newcolumntype{C}[1]{>{\centering\arraybackslash}p{#1}}
\newcolumntype{R}[1]{>{\raggedleft\arraybackslash}p{#1}}

\newcounter{myenum}
 {\end{list}}

\begin{document}
\mainmatter              
\title{Taking Recoveries to Task: Recovery-Driven Development for Recipe-based Robot Tasks}
\titlerunning{Taking Recoveries to Task}  
%
\author{Siddhartha Banerjee\thanks{Indicates equal contribution} \and Angel Daruna$^*$ \and David Kent$^*$ \and Weiyu Liu$^*$ \and Jonathan Balloch \and Abhinav Jain \and Akshay Krishnan \and Muhammad Asif Rana \and Harish Ravichandar \and Binit Shah \and Nithin Shrivatsav \and Sonia Chernova}

\authorrunning{Banerjee et al.} 
%
\tocauthor{Siddhartha Banerjee, David Kent, Angel Daruna, Weiyu Liu, Muhammad Asif Rana, Harish Ravichandar, Abhinav Jain, Akshay Krishnan, Binit Shah, Nithin Shrivastav, and Sonia Chernova}
\institute{Georgia Institute of Technology, Atlanta GA 30332, USA,\\
Corresponding authors: \email{\{siddhartha.banerjee,dekent,adaruna3,wliu88\}@gatech.edu}}

\maketitle              

\begin{abstract}
Robot task execution when situated in real-world environments is fragile.  As such, robot architectures must rely on robust error recovery, adding non-trivial complexity to highly-complex robot systems.  To handle this complexity in development, we introduce Recovery-Driven Development (RDD), an iterative task scripting process that facilitates rapid task and recovery development by leveraging hierarchical specification, separation of nominal task and recovery development, and situated testing.  We validate our approach with our challenge-winning mobile manipulator software architecture developed using RDD for the FetchIt! Challenge at the IEEE 2019 International Conference on Robotics and Automation.  We attribute the success of our system to the level of robustness achieved using RDD, and conclude with lessons learned for developing such systems.
\end{abstract}

\vspace{-0.7cm}
\section{Introduction}
\label{sec:introduction}
\vspace{-0.3cm}

Robot execution is fragile and often overfits to the development test bed~\cite{Atkeson2015,DRC-what-happened}. As such, robust robot architectures must rely on recovery behaviors in order to maintain autonomy when assumptions are violated~\cite{Srinivasa2012}. Recovery during robot tasks is non-trivial, however, as resetting to a known state can be difficult~\cite{Lemaignan2015} and knowing where to resume execution can be context dependent~\cite{Bohren2010}. In addition, unforeseen faults create ambiguity in recovery strategies.

In this work, we address the development of robust recovery for recipe-based tasks---a class of robot tasks that dictate a pre-specified sequence of steps to accomplish a goal. Such tasks include common mobile manipulation tasks in unstructured environments, such as kit packing, machine assembly, table setting, or food preparation. Even for seemingly straightforward recipe-based tasks, the many interactions with the environment, and between robot components, lead to faults that are difficult to identify \textit{a priori}~\cite{Guiochet2017}, often resulting in systems that are inflexible or not robust to failures.

State machines~\cite{Bohren2010}, hybrid automata~\cite{eppner2016lessons}, and planning approaches~\cite{Beetz2010} are common methods of sequencing robot execution that can be made robust to failures. However, robustness is often achieved at the cost of a complexity explosion in the task sequence specification or a loss of interpretability of the task recipe. Crucially, the increased complexity and the lack of interpretability negatively impact the iterative development of the main task and recovery processes, both of which are necessary in the face of potentially innumerable failure conditions.

We therefore propose \textit{Recovery-Driven Development} (RDD), a development process for recipe-based tasks couched in agile methodology.  The key tenet of RDD is the separation of nominal task specification from recovery behavior definition.  Another guiding principle of RDD is the support of hierarchical task specification, which both allows for re-use in the task recipe and provides higher-level context to recovery behavior selection. As such, the RDD methodology enables system developers to easily explore aspects of a robot's system design, such as those identified by Eppner et al.~~\cite{eppner2016lessons}---assumptions, generality, modularity, etc.

We define RDD as a 2-pronged iterative approach to developing robust task execution, in which designers can move back and forth between both prongs without risk of one phase interfering with the other:
\begin{enumerate}
\item \textbf{Specification} (Section~\ref{sec:task-executor}): scripting a hierarchical task sequence incrementally from a task recipe, using strong assumptions
\item \textbf{Refinement} (Section~\ref{sec:task-monitor}): developing recovery behaviors by executing a situated task, noting a fault, specifying new recoveries, and repeating
\end{enumerate}
The result is a development methodology that supports rapid task and recovery prototyping, without a noticeable loss in the robot's robustness when deployed.

In this work, we present our task execution and monitoring system as an example framework designed to enable RDD for recipe-based robot tasks\footnote{\url{https://github.com/GT-RAIL/derail-fetchit-public/tree/master/task_execution}}. 
We validate both the task system and the RDD workflow with our team's winning approach to the FetchIt! Challenge at the IEEE 2019 International Conference on Robotics and Automation (ICRA), in which our success was achieved mainly due to the robustness afforded to our system from the RDD methodology.  Additionally, we provide details and open-source code for our complete system developed for the FetchIt! Challenge as a concrete example of a complex mobile manipulation system developed using RDD.  We conclude with a discussion of lessons learned for the fast and robust development of recipe-based robot tasks.

\vspace{-0.5cm}
\section{Related Work}
\label{sec:related-work}
\vspace{-0.3cm}

Designing a robot's software is often application-dependent, requiring tradeoffs between multiple approaches~\cite{Kortenkamp2016}. In this section we enumerate common design choices that emerge across applications for robot architectures, situate our task execution framework within the design practices, and motivate the development of our approach.

Robot architectures are generally three-tiered with the following levels~\cite{Kortenkamp2016}:
\begin{itemize}
	\item A \textit{behavioral} level for highly reactive and highly situated robot execution. Modules at this level, sometimes termed skills, have a tight perception-action loop and are the focus of much research. 
	\item An \textit{executive} level that bridges low-level tasks (skills) and high-level tasks (goals). The executive is responsible for sequencing skills, monitoring execution, and handling exceptions. 
	\item A \textit{planning} level responsible for tasking the executive level with goals to achieve based on future objectives, robot constraints, environmental situations, etc. 
\end{itemize}
Our primary contribution is in enabling the executive level to support an RDD workflow, and as such the remainder of this section examines executive level design and recovery.  We discuss a behavioral implementation of mobile manipulation in Section \ref{sec:system} to provide context for our executive implementation.  We also note that planning-level requirements are minimal for autonomous recipe-based tasks, although we return to this assumption at the end of Section~\ref{sec:discussion}.

\vspace{-0.3cm}
\subsection{Executive Level Design}
\label{sec:related-executive}
\vspace{-0.3cm}
The most informative consideration in executive level design is how dynamic or static the task should be. A task can be dynamic due to environments with uncontrolled agents such as humans~\cite{Berenz2018} or competing objectives~\cite{Szafir2017}. There are four paradigms to behavior sequencing at the executive level, with differing levels of support for dynamic tasks:

	\begin{itemize}
	\item \textit{Agent-based control} partitions control into separate, synchronized agents that maintain consistency with the global robot objective. This works well for dynamic environments, and was implemented through behavior trees in Playful~\cite{Berenz2018} and resource agents in ROAR~\cite{Degroote2011}. However, debugging and reasoning about the interactions between agents can be difficult.
	
	\item \textit{Planning} is a principled manner of sequencing skills in dynamic environments, and was implemented by CRAM~\cite{Beetz2010}. However, when designers know the exact sequence of skills they want, as in recipe-based tasks, the design process for planning can be non-intuitive or even counter-productive~\cite{Bohren2010}.
	
	\item \textit{Finite State Machines} retain some of the autonomy in sequence specification provided by planning, and also allow system designers to explicitly specify state transitions a priori based on expected sub-task outcomes. Additionally, state machines support model verification and composition for incremental construction of complex behaviors~\cite{klotzbucher2010orocos,Bohren2010}. However, state machines, and the related method of hybrid automata~\cite{eppner2016lessons}, suffer from an explosion of transitions as the number of skills or the task complexity grows. 
	
	\item \textit{Scripting} allows for the compositionality of state machines with the simple declaration, rather than programming, of robot behavior~\cite{Srinivasa2012}. Additionally, scripting provides easier error recovery to handle exceptions at the executive level than state machines. However, scripting puts the burden of sequence specification on the designer, raising scalability issues, especially for multi-objective tasks~\cite{Bohren2010}.
	\end{itemize}

In this work, we focus on relatively static environments and recipe-based tasks that can be decomposed into subtasks. In order to facilitate the rapid prototyping and incremental inclusion of error recoveries inherent to RDD, we sequence behaviors through hierarchical scripting. We address scripting's scalability issues by separating task specification and recovery. 

\vspace{-0.3cm}
\subsection{Reactivity and Recovery}
\label{sec:related-recoveries}
\vspace{-0.2cm}
A robust robot executive level must include failover mechanisms that maintain autonomy when behavior design assumptions are not satisfied~\cite{Srinivasa2012}. Therefore, recovery systems must address many challenges, including determining how to reset to a known state~\cite{Lemaignan2015}, handling context dependent execution resumption~\cite{Bohren2010}, and deciding on recovery strategies for unforeseen faults. A common recovery strategy for resetting to a known state is to re-attempt the entire task, as in \cite{eppner2016lessons}, although such approaches are less reactive to failures.

Planning approaches maintain reactivity by recovering from seen and unforeseen failures by replanning~\cite{Beetz2010}. Further, plans provide theoretical guarantees on robustness to unforeseen execution failures~\cite{Kress-Gazit2009,eppner2016lessons}. Prior works have treated recovery as a planning problem  with the goal of reaching any state where a diagnosed fault does not exist~\cite{Zaman2013}. However, as noted earlier, planning approaches can be difficult to iteratively develop, or can result in not-easily interpretable task specifications.  Our approach avoids planning at the executive layer in favor of scripting, to facilitate rapid and highly-interpretable behavior development.

In the absence of planning, reactively resetting to a good known state can be accomplished through fault forecasting, such as FMECA, which reasons about expected faults and the explicit recovery steps to address them~\cite{Crestani2015}. However, such approaches are time consuming and not guaranteed to find all faults~\cite{Guiochet2017}. We instead take an empirical approach to fault discovery through situated task execution, exploiting the inherent structure of recipe-based tasks, allowing for pre-scripted recovery to intermediate task steps.

\vspace{-0.3cm}
\section{System Overview}
\label{sec:system}
\vspace{-0.3cm}

\begin{figure}[t]
\centering
\includegraphics[width=1.0\textwidth]{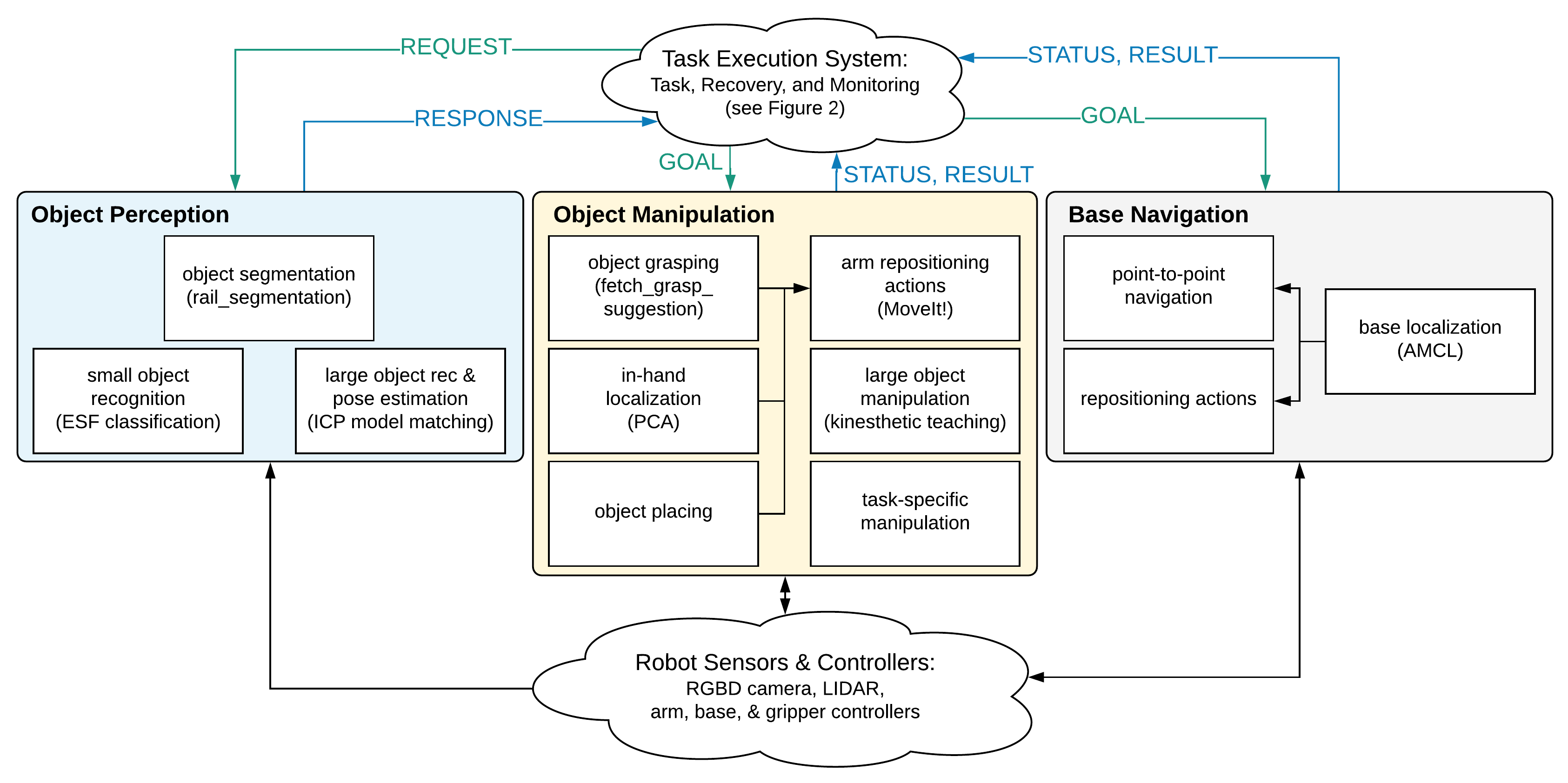}
\vspace{-0.5cm}
\caption{Mobile manipulation system overview. Arrows denote ROS information flow, through publishers, subscribers, services, and actionlib.}
\label{fig:ros-system-diagram}
\vspace{-0.7cm}
\end{figure}

Before presenting our task execution and recovery system, we first describe the behavioral level of our general mobile manipulation architecture, to serve two purposes: (1) to share our open-source challenge-winning mobile manipulation system developed to support RDD, and (2) to establish a task context and a set of robot capabilities that we will refer to throughout the paper, grounding our discussion of RDD's benefits and drawbacks in a fully-realized robot system.  The architecture consists of a set of independent mobile manipulation modules, implemented using the Robot Operating System (ROS) \cite{quigley2009ros}, 
shown in Figure~\ref{fig:ros-system-diagram}. Object perception modules are implemented as ROS service servers, and object manipulation and base navigation modules are implemented as \texttt{actionlib}\footnote{\scriptsize http://wiki.ros.org/actionlib} servers.  Each independent module can be called by the task executor, and provides feedback to the task executor and task monitor\footnote{Each module must necessarily provide feedback on its own faults so that the executive level can make relevant recovery decisions.}. The modules consist of the following capabilities:

\textit{Object Perception.} Our perception modules implement a perception pipeline for RGBD sensor data, using the Point Cloud Library (PCL) \cite{rusu2011point}.  The object segmentation module uses the \texttt{rail\_segmentation}\footnote{\scriptsize \url{http://wiki.ros.org/rail_segmentation}} package to identify point cloud clusters-of-interest through table surface detection and Euclidean distance clustering.  We divide our object recognition approaches between large and small objects.  For large objects, we perform model matching using the \texttt{rail\_mesh\_icp}\footnote{\scriptsize \url{http://wiki.ros.org/rail_mesh_icp}} package that uses an Iterative Closest Point (ICP) PCL pipeline, which also provides object pose detection.  For small object recognition, we train an SVM classifier over Ensemble of Shape Functions (ESF) descriptors \cite{wohlkinger2011ensemble}.  We do not need to perform pose estimation for small objects due to our object grasping approach, described below.

\textit{Object Manipulation.} Most of our manipulation modules make use of MoveIt! to perform arm planning to either joint goals or end-effector pose goals using OMPL's \texttt{RRTConnect} motion planner \cite{chitta2012moveit}.  This includes both general arm repositioning actions, which the task executor can call directly (e.g. to move the arm out of the way of the camera), and execution actions, called by other object manipulation modules.  Object grasping calculates antipodal grasps over an object point cloud using the \texttt{agile\_grasp} package \cite{ten2018using}, which are then ordered and executed using pairwise ranking through \texttt{fetch\_grasp\_suggestion} \cite{kent2018adaptive}.  As objects can shift during the grasping process, we perform post-grasp pose detection using the in-hand localization module, which identifies the object point cloud by performing background subtraction on the robot's gripper, and calculates the object's pose based on its principal axes determined by Principal Component Analysis (PCA).  Given a known object pose and a desired place location, object placing calculates and executes a pose goal for placing an object that ensures the gripper fingers and palm are out of the way of the object's fall trajectory.

We also include some manipulation modules that do not use MoveIt!, due to the limitations of sampling-based motion planning.   For large object manipulation, such as lifting and placing kits of objects, we include a kinesthetic teaching module \cite{akgun2012trajectories}.  This allows system designers to record and play back arm trajectories, either in full or as a set of waypoints.
Additionally, we include task-specific manipulation actions to implement specific manipulation skills such as raising and lowering objects, using a Cartesian end-effector controller\footnote{\scriptsize Available at \url{https://github.com/GT-RAIL/fetch_simple_linear_controller}}, and peg-in-hole insertion, using a controller with end-effector pose and joint effort as feedback.

\textit{Base Navigation.} LIDAR-based localization uses AMCL provided by ROS's \texttt{nav\_stack}\footnote{\scriptsize \url{http://wiki.ros.org/navigation}} to localize the base with respect to a pre-collected 2D occupancy grid of the environment.  Navigation is primarily done using point-to-point navigation between waypoints on the map, executed using a PID controller\footnote{In complex environments, \texttt{nav\_stack}'s global and local planners can be used instead.}.  We also include local repositioning actions, which implement short movement primitives such as backing up from a table. The repositioning actions are implemented using a PID controller with gains tuned for shorter, more precise base goals.

\medskip

With each module implemented, the navigation, perception, and manipulation actions can be sequenced in a robust manner to complete mobile manipulation tasks by the task execution system described in the next section.


\vspace{-0.3cm}
\section{Task Execution and Recovery}
\label{sec:executive}
\vspace{-0.3cm}

\begin{figure}[t]
\centering
\includegraphics[width=0.75\textwidth]{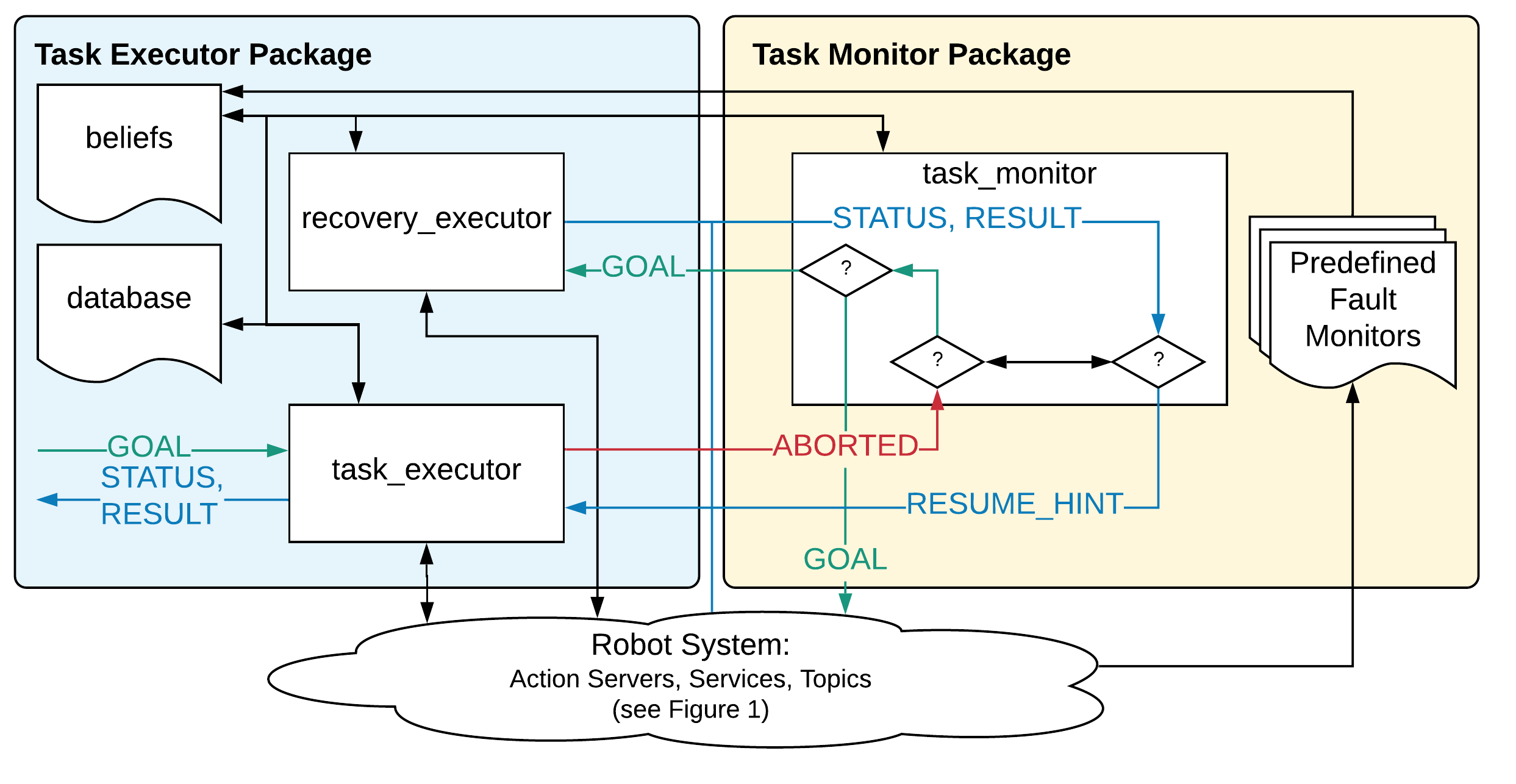}
\vspace{-0.3cm}
\caption{Overview of the two packages in our the executive level.  Arrows denote ROS information flow, through publishers, subscribers, services, and actionlib.}
\label{fig:task-package-diagram}
\vspace{-0.5cm}
\end{figure}

In this section, we describe our executive level, which consists of two packages seen in Figure~\ref{fig:task-package-diagram}: the \textit{task executor} and the \textit{task monitor}\footnote{\scriptsize Stand-alone packages under development at {\url{https://github.com/GT-RAIL/assistance_arbitration}}}. The task executor (Section~\ref{sec:task-executor}) contains scaffolding to specify and incrementally develop a main task recipe. The task monitor (Section~\ref{sec:task-monitor}) contains the utilities necessary recover from general failures during task execution.

\vspace{-0.3cm}
\subsection{RDD Specification: The Task Executor}
\label{sec:task-executor}
\vspace{-0.3cm}



During the \textit{Specification} phase of RDD, developers translate the nominal behaviour of the robot executing a task recipe into a script. Crucially, developers should fulfill two objectives in this phase: (1) declare robot behavior under the strong assumption of perfect robustness in execution, and (2) provide structure to the specified script so that in the event of an error, it is easy to garner error context as well as resume execution once the error is resolved.  The task executor package facillitates meeting such objectives.

Specifically, the task executor provides the following utilities to aid in rapid task prototyping and testing: 
\begin{itemize}
\item A Python-based abstraction for specifying semantically meaningful interfaces to the robot's behavior layer, to form sequenceable primitive actions.
\item A custom domain-specific language using YAML syntax for scripting recipe-based tasks, with a view towards facilitating hierarchical task declaration for code modularity and reuse.
\item A consistent API to tasks and actions to facilitate testing in isolation and to enable easy invocation from other system components.
\item A database to provide a common knowledge-base of task relevant information to all tasks and primitive behaviors.
\item An automatically populated belief system encapsulating pertinent robot, environment, and task states, to provide additional context during recovery.
\end{itemize}
The following sections provide additional details on the above utilities.

\vspace{-0.3cm}
\subsubsection{Actions}
\label{sec:task-actions}

Primitive actions are specified as Python objects derived from a common abstract class. They are implemented either as a client to individual robot components, such as to point-to-point navigation, or as a client to semantic groupings of robot components, such as to the grasp calculation packages.

\vspace{-0.3cm}
\subsubsection{Tasks}
\label{sec:task-def}

Tasks manifest as a Python class derived from the same abstract class as \textit{actions}, but whose execution is specified using a custom domain-specific language, which uses YAML syntax, to allow loading and reloading of tasks from the ROS parameter server. The language allows tasks to:

\begin{itemize}
\item reuse other tasks for the creation of complex task hierarchies
\item accept parameters for adaptation and compositionality in task specification
\item create, maintain, and manipulate local variables for data transfer between actions and for adaptation to environmental or execution conditions
\item utilize rudimentary control flow through conditional statements and loops, aiding in concise task specification
\end{itemize}

\begin{figure}[t]
\begin{minipage}[b]{.39\textwidth}
\begin{listing}[H]
\begin{minted}[fontsize=\tiny]{yaml}
(task name):
  [params: [...]]
  [var: [...]]
  steps:
  - (action | task | op | choice | loop) : (name)
    [params: {...}]
    [var: [...]]
  
  [...]
\end{minted}
\caption{Task Syntax}
\label{lst:task-syntax}
\end{listing}
\end{minipage}
\noindent\begin{minipage}[b]{.6\textwidth}
\begin{listing}[H]
\noindent\begin{minipage}[b]{.49\textwidth}
\begin{minted}[fontsize=\tiny]{yaml}
detect_schunk_pose_task:
  params:
  - look_location

  var:
  - chuck_approach_pose

  steps:
  - action: look
    params:
      pose: params.look_location

  - action: detect_schunk
    var:
    - chuck_approach_pose
\end{minted}
\end{minipage}
\noindent\begin{minipage}[b]{.5\textwidth}
\begin{minted}[fontsize=\tiny]{yaml}
pick_task:
  params: [object_idx, grasps, object_key]
  var: [grasped]
  steps:
  - action: pick
    params:
      object_idx: params.object_idx
      grasps: params.grasps
      object_key: params.object_key

  - action: verify_grasp
    params:
      abort_on_false: false
    var:
    - grasped
\end{minted}
\end{minipage}
\caption{Example Tasks}
\label{lst:task-example}
\end{listing}
\end{minipage}
\vspace{-0.7cm}
\end{figure}

We present the formal task syntax in Listing~\ref{lst:task-syntax}, with example tasks used for large object pose estimation and for object picking at the FetchIt! Challenge (Section~\ref{sec:fetchit-challenge}) shown in Listing~\ref{lst:task-example}. A task is defined as a dictionary entry with the required key of \texttt{steps}, which defines a list of the steps in the task, and the optional keys of \texttt{params} and \texttt{var}, which define the task inputs and outputs. Each \texttt{step} in the task is named and can be one of five types: (1) \texttt{action}, invoking a primitive action, (2) \texttt{task}, invoking another task, (3) \texttt{op}, invoking a simple Python function for rudimentary data manipulation, (4) \texttt{choice}, to evaluate a boolean expression for control flow, and (5) \texttt{loop}, to loop while a boolean expression is \texttt{true}. All steps accept a dictionary setting values for their \texttt{params}, and return a list of \texttt{var} values that become local variables in the parent task\footnote{For more details, see the README in our Github repository.}.

\vspace{-0.3cm}
\subsubsection{Consistent API}
\label{sec:task-api}

Tasks and actions have a consistent API, which mimics that of ROS's \texttt{actionlib} interface. This consistent API enables (1) the use of JSON to specify inputs and outputs to tasks and actions easily in order to test them in isolation, and (2) the invocation of individual tasks from other ROS nodes, such as the recovery system, through the \texttt{actionlib} interface when required.

\vspace{-0.3cm}
\subsubsection{Database}
\label{sec:task-database}

Recipe-based tasks can be parameterized by semantically meaningful task variables which are then grounded to different values for particular environments or tasks. Example task variables are locations, robot poses, objects, or other real-world entities. The database is a YAML dictionary loaded into the ROS parameter server that provides a single source of truth for grounding all relevant task variables. Rapid environment adaptation is readily facilitated by modifying the values associated with known keys in the database definition.

\vspace{-0.3cm}
\subsubsection{Beliefs}
\label{sec:task-beliefs}

Beliefs are included in the task executive layer for two reasons: (1) to provide context to recovery mechanisms in the event of a failure, and (2) to provide updates to a higher level planner, should one exist, about relevant states of the task, the robot, or the environment. For instance, the expected and actual state may become mismatched: transient localization and navigation errors during point-to-point navigation might compound to leave the robot at a location outside an expected tolerance for manipulation actions. Background monitors on the robot's location can indicate a mismatch, and in turn the recovery system can use this information to reposition the robot.

\vspace{-0.3cm}
\subsubsection{Discussion}
\label{sec:task-discussion}

The task executor package is optimized to facilitate rapid specification and testing of recipe-based tasks: developed task scripts are deterministic, easy to specify, interpretable, and readily allow the testing of components in isolation. Additionally, the scripting approach to task specification provides the implicit benefits of straightforward state tracking and an efficiency in task execution borne from overestimating the robustness of the robot's behaviors. Indeed, we do not check for most violations to the operating conditions of our primitive behaviors until they report a failure.

We note that the determinism of our scripting approach and overestimation of the robustness of our behaviors leaves us susceptible to violations in assumptions of the environmental state (a susceptibility that reactive sequencing approaches do not share). However, instead of complicating the task scripts and in turn slowing down task specification, our RDD methodology relies on the incremental recovery development to achieve robustness and reactivity.

\vspace{-0.3cm}
\subsection{RDD Refinement: The Task Monitor}
\label{sec:task-monitor}
\vspace{-0.2cm}

The primary objective of system development during the \textit{Refinement} phase of RDD is to rapidly incorporate diverse recovery strategies for a specified task recipe in order to incrementally improve its robustness. As such, it involves addressing four challenges (mentioned in Section~\ref{sec:related-work}):
\begin{enumerate}
\item resolving ambiguity in the recovery policy for unforeseen faults
\item taking actions to reset to a known state in the event of a fault
\item deciding how to resume execution once a fault is addressed
\item trying diverse strategies when recovering from a repeated fault
\end{enumerate}
The task monitor package, which provides execution monitoring and error recovery to the task executor, is designed to address each of the above challenges.


\vspace{-0.3cm}
\subsubsection{Handling Unseen Errors}
\label{sec:monitor-unseens}

In the event of an unseen error during development\footnote{The system can detect unseen errors in three ways: (1) the behavior level can propagate reported faults (i.e. action servers aborting, nodes crashing, etc.), (2) recipe steps can explicitly check for expected errors, or, in the case of unexpected errors, (3) the developer can stop system execution and write a new error detection module.}, the monitor immediately exits from the task, displaying the entire context of the error in a consistent manner and logging all possible causes. Developers can then inspect the logs to create a tailored (set of) recovery mechanism(s) for such errors, thereby making them ``known'' errors during future failures. In practice, we quickly accumulate a list of errors that our developed recovery strategies know to address.

During deployment, unseen errors can be dealt with under a domain-dependent context-relevant policy of always exit, always retry, or some combination thereof.

\vspace{-0.2cm}
\subsubsection{Taking Actions}
\label{sec:monitor-actions}

Rapidly developing and testing actions to take in the event of a particular failure requires the presence of (1) a means of determining the diagnosis of an error, and (2) an easy mechanism to invoke actions or subsets of actions. The monitor and task executor are designed to facilitate both.

\begin{figure}[t]
\centering
\includegraphics[width=\textwidth]{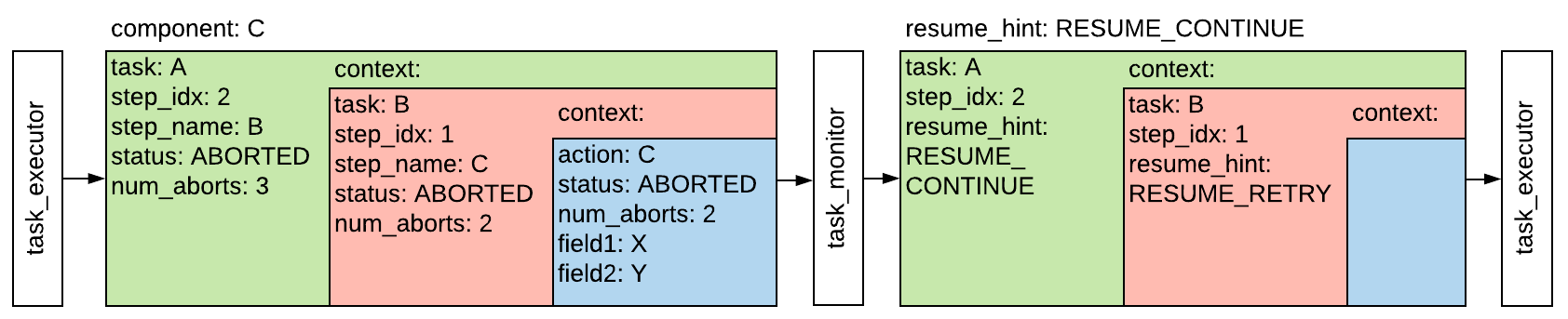}
\caption{Metadata passed between the task executor and the task monitor, used to facilitate error diagnosis and task resumption after fault resolution.}
\label{fig:context-dictionaries}
\vspace{-0.5cm}
\end{figure}

The structure and compositional design of our main task recipe aids in fault diagnosis, given the current task state. Specifically, in the event of an error, tasks in the task executor provide a consistent context of their state in a recursive dictionary containing all tasks in a task hierarchy until the primitive action, and primitive actions can also provide error context through custom data fields, as shown in Figure~\ref{fig:context-dictionaries}. Further, the task executor's beliefs (Section \ref{sec:task-beliefs}) can additionally inform error recovery.

When an error is diagnosed, the consistent API for invoking tasks and actions facilitates the monitor in resolving the problem. The monitor uses a redundant instance of the task executor, called the recovery executor (seen in Figure~\ref{fig:task-package-diagram}), to execute simple task recipes for recovery.

\vspace{-0.3cm}
\subsubsection{Resuming}
\label{sec:monitor-resume}

We have identified five strategies for resuming task execution that have applied to the errors we have encountered:

\begin{enumerate}
\item \texttt{RESUME\_NONE}: stop executing the task.
\item \texttt{RESUME\_CONTINUE}: resume task execution from the failed step.
\item \texttt{RESUME\_RETRY}: restart a subtask, or the whole task; useful if, for example, the environment changed during recovery and thus perception must be rerun.
\item \texttt{RESUME\_NEXT}: resume execution at the next step; useful if the recovery process accomplishes the failed step.
\item \texttt{RESUME\_PREVIOUS}: resume execution at the previous step; useful if failure assumptions change, but the entire subtask does not need to be restarted.
\end{enumerate}
To enable full flexibility in the recovery mechanism on how tasks\footnote{Resuming execution from arbitrary stopping points in primitive actions is hard~\cite{Lemaignan2015}, but depending on the implementation of the robot system, might be unnecessary.} are resumed, any of the tasks in a hierarchy can be resumed using any of the above five strategies. An example of the context for task resumption is shown in Figure~\ref{fig:context-dictionaries}.

\vspace{-0.3cm}
\subsubsection{Recovery Diversity}
\label{sec:monitor-diversity}

Due to the larger context of some errors, the same recovery actions taken during the same fault diagnosis can fail: for example, recalculating grasps on a small object when sampling-based arm motion planning fails to pick it up may be insufficient due to an arm workspace limitation, and instead the error should be resolved by repositioning the robot base or by moving the arm to a different start configuration. As such, the context dictionaries included for diagnosis and resumption support development of diverse recovery strategies for the same faults, based on factors such as task hierarchy location, primitive action failure count, or hierarchical task failure counts.

\vspace{-0.3cm}
\subsubsection{Discussion}
\label{sec:monitor-discussion}

The philosophy behind RDD's \textit{Refinement} phase, i.e. incremental and independent recovery development, necessitated a recovery system that is deterministic, easy to specify, interpretable, and readily allows testing of individual recoveries in isolation. Although the current version of our implemented recovery system is not robust to failures during the recovery process, such robustness can either be added in a future iteration of our system, or can be left to the purview of a higher-level planner in the robot system. Finally, we note that our current rule-based system of recoveries does not easily lend itself to analysis or verification, but such a feature can be integrated in the near future. In the meanwhile, the easy testability of individual recoveries mitigates the lack of verification in the system.

\vspace{-0.3cm}
\section{Validation}
\label{sec:validation}
\vspace{-0.3cm}

Our task execution approach formed our executive level for the FetchIt! Challenge at ICRA 2019.  The challenge's goal was to advance autonomy and robustness by using a mobile manipulator to perform an industrial kit assembly task in an unstructured environment.  As such, the challenge was a good opportunity to validate our system and development methodology in a real-world, time-sensitive scenario.  We provide a brief description of the FetchIt! Challenge, followed by quantitative and qualitative observations of the RDD workflow and the recovery mechanisms we developed for our competition-winning robust task executor.

As further context, and to provide a continuous example of recoveries over a 45-minute autonomous task, we provide a video of our final competition run\footnote{ \url{https://youtu.be/G_ur71h4CNQ}}.

\vspace{-0.3cm}
\subsection{FetchIt! Challenge Overview}
\label{sec:fetchit-challenge}
\vspace{-0.1cm}

\begin{figure}[t]
	\centering
	\subfloat[Fetch Robot]{
		\includegraphics[width=0.195\textwidth]{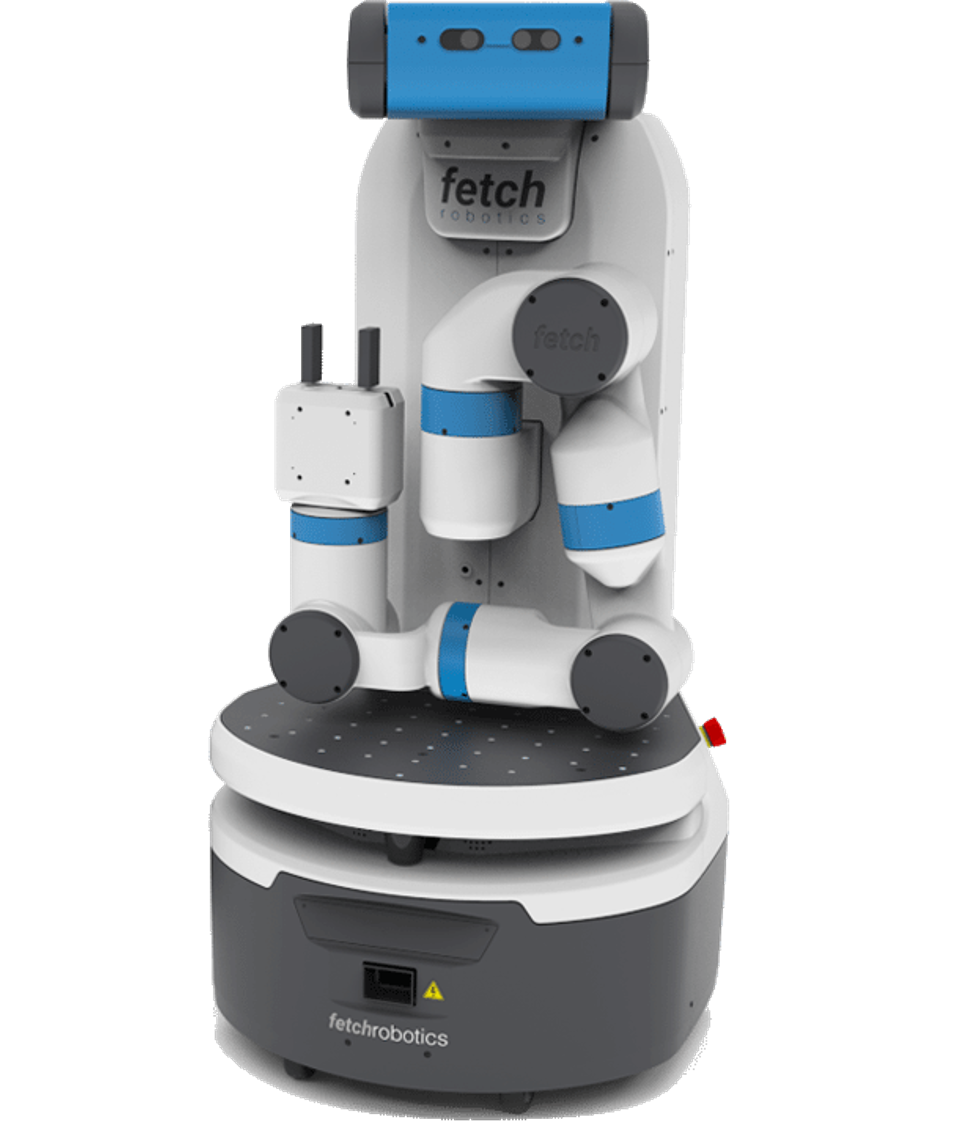}
		\label{fig:fetchit-fetch}
	}
	\subfloat[Challenge Arena]{
		\includegraphics[width=0.25\textwidth]{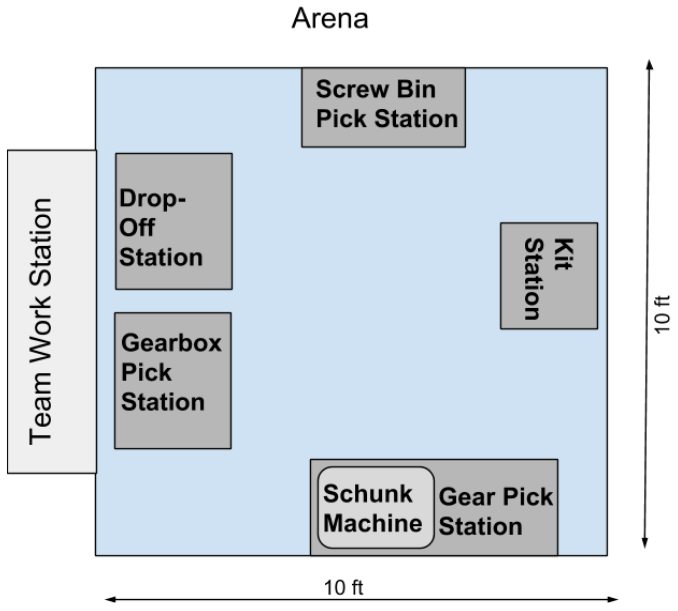}
		\label{fig:fetchit-arena}
	}
	\subfloat[Assembled Kit]{
		\includegraphics[width=0.225\textwidth]{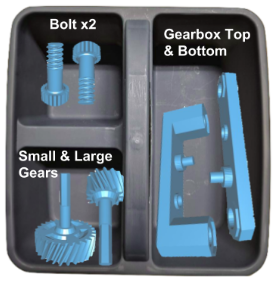}
		\label{fig:fetchit-kit}
	}
	\subfloat[Schunk Machine]{
		\includegraphics[width=0.25\textwidth]{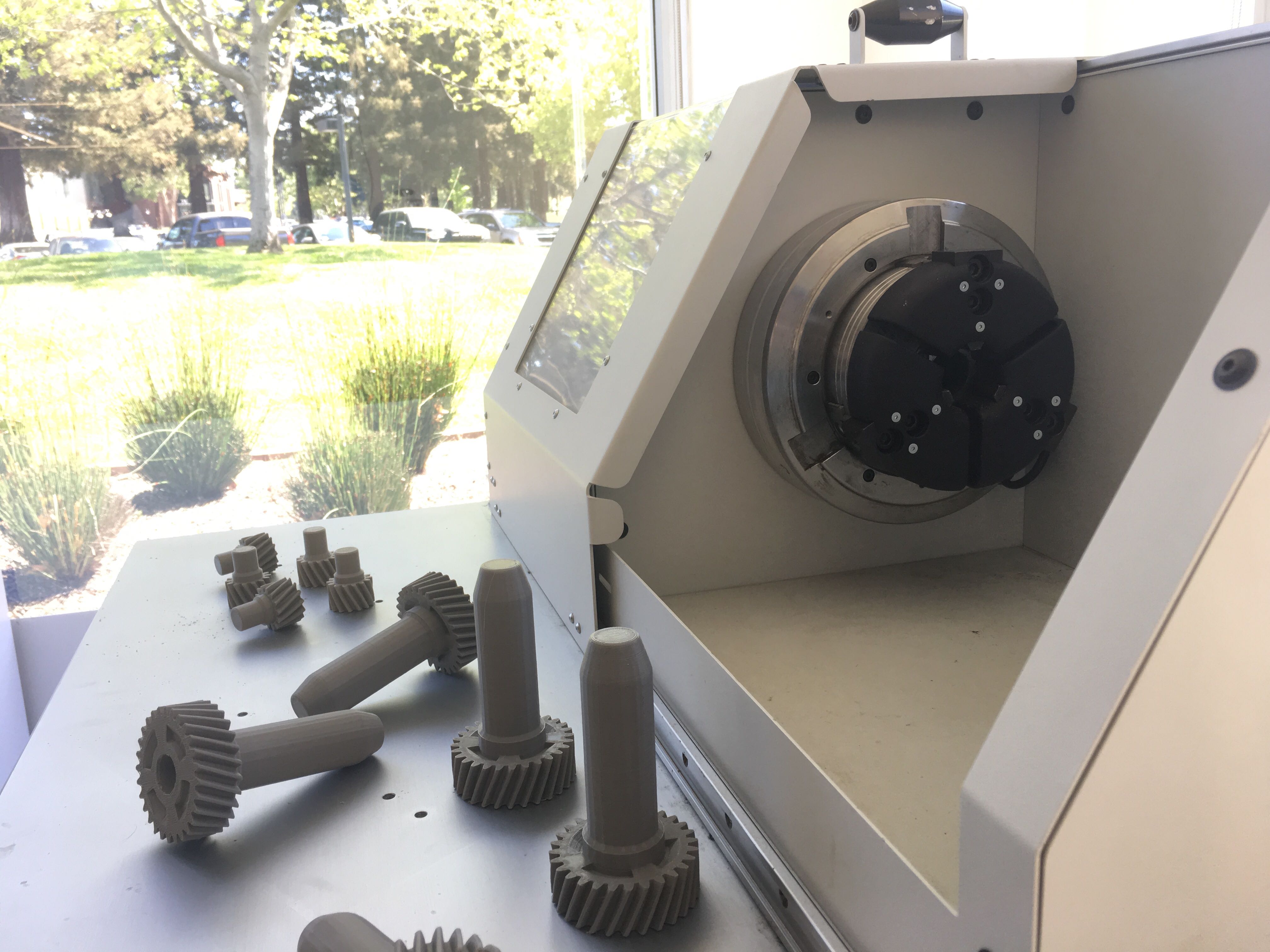}
		\label{fig:fetchit-schunk}
	}
	\caption{FetchIt! challenge hardware and specifications.}
	\label{fig:fetchit-challenge}
\vspace{-0.5cm}
\end{figure}

The FetchIt! Challenge was a mobile-manipulation challenge focused on autonomously completing combined manipulation, perception, and navigation tasks on a mobile-manipulator platform\footnote{\scriptsize \url{https://opensource.fetchrobotics.com/assets/Rulebook2019.pdf}}.  Specifically, the goal of the competition was to have a Fetch mobile manipulator~\cite{wise2016fetch}, equipped with an RGBD camera and a 2D LIDAR (Figure~\ref{fig:fetchit-challenge}a), autonomously assemble kits (Figure~\ref{fig:fetchit-challenge}c). In order to assemble each kit, the Fetch had to navigate a challenge arena (Figure~\ref{fig:fetchit-challenge}b), perceiving and picking the various parts from tabletops and bins. In addition to pick and place, the Fetch had to operate machinery in the arena via physical manipulation and wireless interfaces as part of the assembly process. For instance, it had to insert the ``Large Gear'' in Figure~\ref{fig:fetchit-challenge}c into a small opening shown in Figure~\ref{fig:fetchit-challenge}d on a simulated milling machine.  Perfectly completed kits (as in Figure~\ref{fig:fetchit-challenge}c) scored 7 points, with no points awarded for incomplete kits (i.e. any parts missing).

The robot was required to run autonomously with no intervention for an allotted time of forty five minutes, completing as many kit assemblies as possible.  The strict scoring requiring fully assembled kits made robust task execution one of the largest challenges of the competition.
\vspace{-0.5cm}

\subsection{Validation of Recovery-Driven Development}
\label{sec:rdd-validation}

\begin{figure}[t]
	\centering
	\subfloat[The top three levels in the hierarchical task tree. Repetitions of  $pick\_place\_object\_in\_kit$ are omitted and denoted with an ellipsis for brevity.]{
		\includegraphics[width=1.0\textwidth]{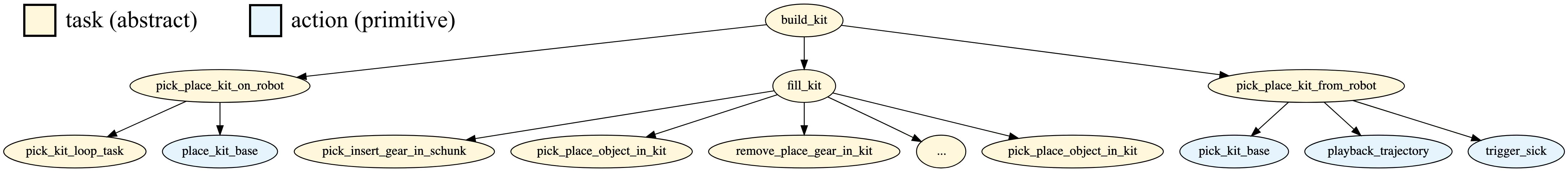}
		\label{fig:high-level-task-flow}
	}
	\vspace{2mm}
	\subfloat[Full expansion of the fifth $pick\_place\_object\_in\_kit$ task. The suffix $n$ for each node indicates the $n^{th}$ invocation of an action or task over the full task tree ($n$ is often higher in practice due to recovery execution). Details of $place\_in\_kit\_task$ are omitted for brevity. $C$ denotes a choice node that denotes conditional execution, and $L$ denotes a loop node.]{
		\includegraphics[width=1.0\textwidth]{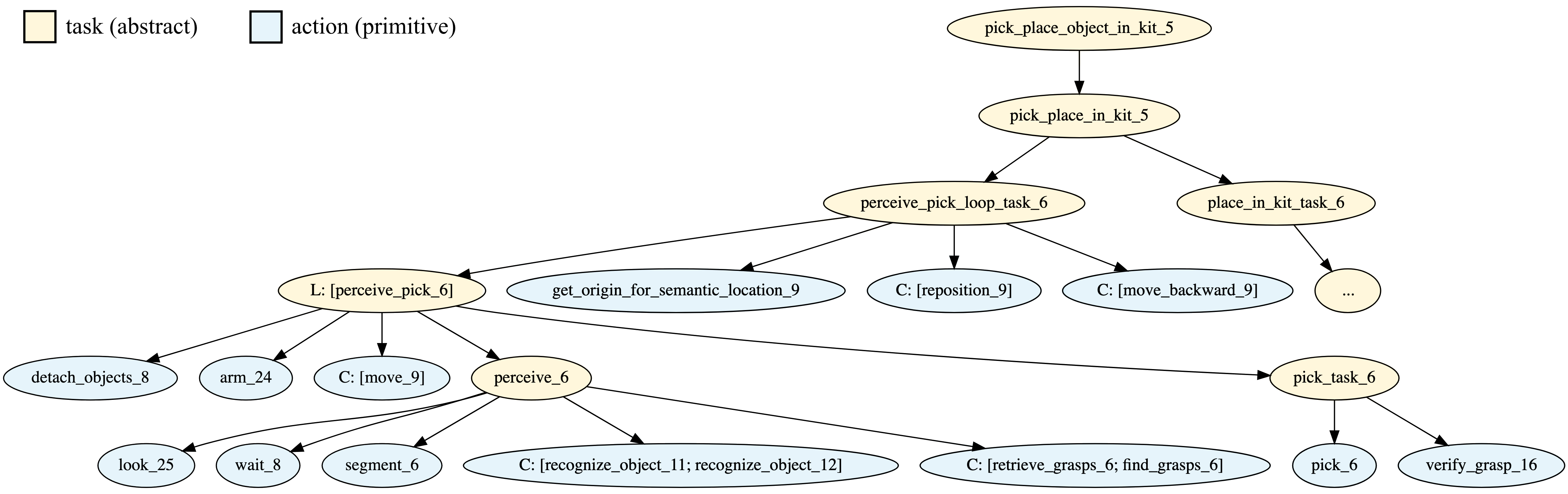}
		\label{fig:low-level-task-flow}
	}
	\caption{Hierarchical task tree for the FetchIt! challenge.}
	\label{fig:task-flow}
\vspace{-0.5cm}
\end{figure}


Figure~\ref{fig:high-level-task-flow} shows the high-level structure of the task implemented for the FetchIt! Challenge---an easy to understand script. Figure~\ref{fig:low-level-task-flow} demonstrates the task complexity, showing an expansion of one of the abstract tasks from Figure~\ref{fig:high-level-task-flow}. This complexity is manageable thanks to hierarchy-enabled action and task reuse--our task script reuses the look action at least 25 times and the perceive task at least 6 times. The modular nature of the task specification simplified the design process and allowed for independent testing of task recipes.


Further, the RDD requirement of separating specification and refinement allowed us to safely develop new recovery behaviors in high-pressure moments between competition runs.   For instance, the simulated milling machine required gear insertion into a smaller hole than we had previously tested with, which caused new faults resulting from false positives in the robot's evaluation of the insertion.  With only 45 minutes to test between runs, we were able to quickly update existing recovery strategies to identify insertion failure and retry the task, without risk of disrupting the previously tested nominal task flow.


\vspace{-0.3cm}
\subsection{Evaluation of Task Robustness}
\vspace{-0.1cm}

The FetchIt! Challenge provided an opportunity to gather quantitative data on our recovery strategies, allowing us to evaluate the error recovery utilities provided by the task monitor (Section~\ref{sec:task-monitor}). We also provide a representative example to qualitatively highlight those utilities.

\vspace{-0.7cm}
\subsubsection{Quantitative Observations}
We provide a breakdown of the recovery strategies we implemented for the FetchIt! Challenge.  In total, we implemented 18 strategies, which often included multiple sub-strategies to handle dynamic execution under differing fault conditions.

\begin{figure}[t]
	\centering
	\setcounter{subfigure}{0}
	\subfloat[Properties of the recovery]{
		\includegraphics[width=0.32\textwidth]{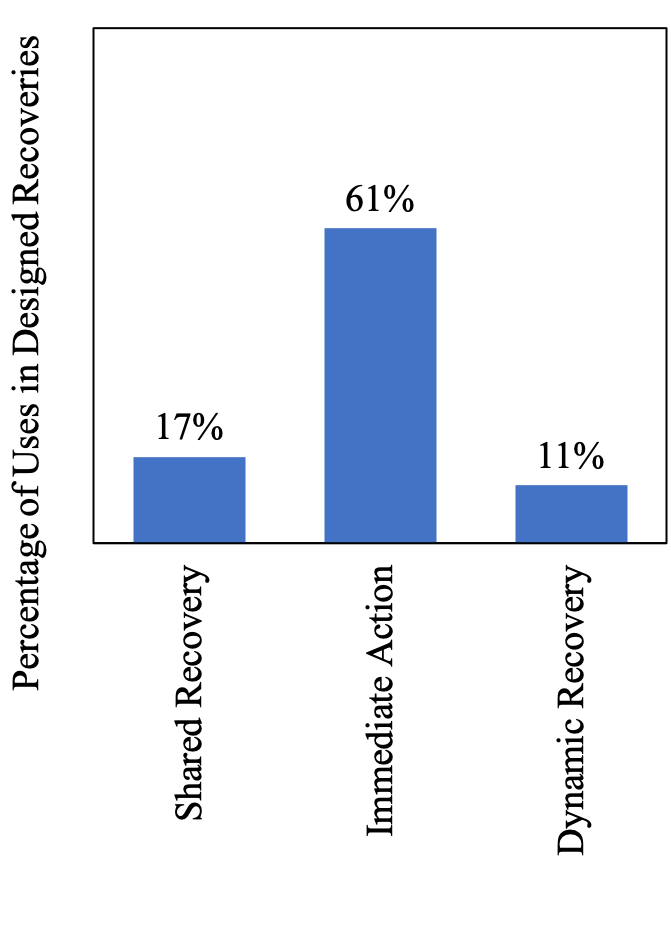}
		\label{fig:recovery-properties}
	}
	\subfloat[Factors used in diagnosis]{
		\includegraphics[width=0.31\textwidth]{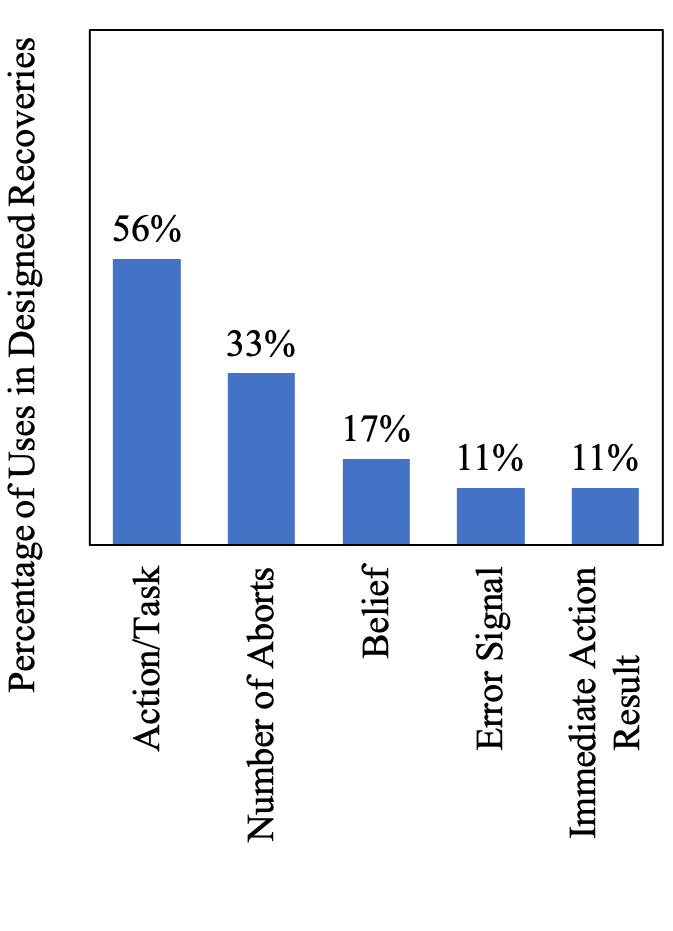}
		\label{fig:fault-diagnosis}
	}
	\subfloat[Resumption strategies]{
		\includegraphics[width=0.3\textwidth]{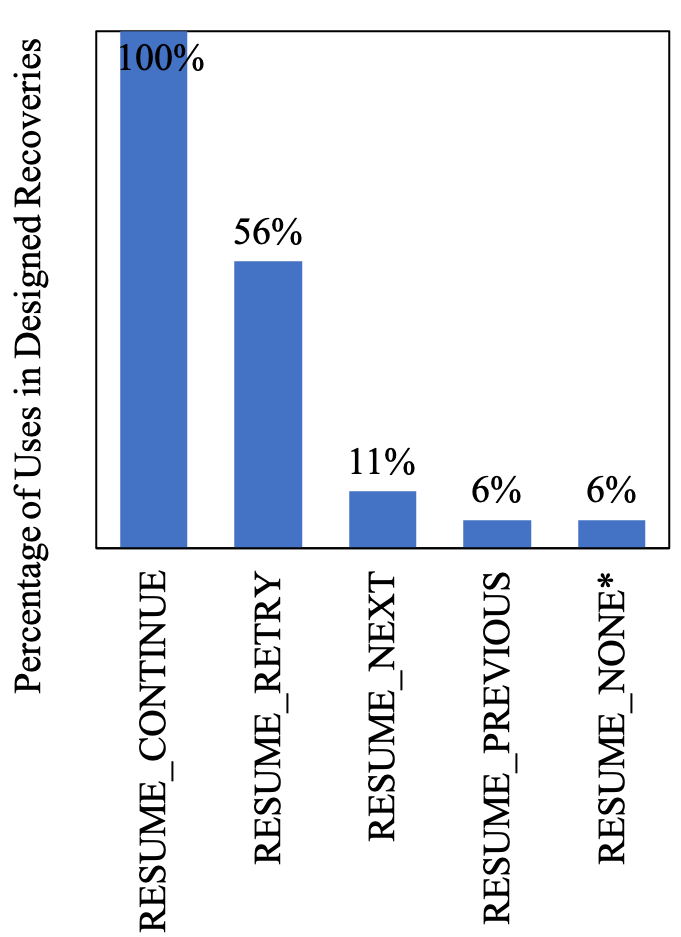}
		\label{fig:resumption-strategy}
	}
	\caption{Percentage of times that recovery uses each utility of the task monitor, for the 18 main recovery strategies designed for FetchIt! Challenge. In (c), note that \texttt{RESUME\_NONE} is also the default strategy for unseen errors.}
	\label{fig:fetchit-challenge}
\vspace{-0.7cm}
\end{figure}

To highlight the value of easy rule specification for recoveries and the ability to act upon a rule-based diagnosis, we define the following three situations:
\begin{enumerate}\vspace{-0.1cm}
  \item \textit{Shared Recovery}: different faults use the same rules for diagnosis and recovery
  \item \textit{Immediate Action}: recovery directly invokes a primitive action
  \item \textit{Dynamic Recovery}: in the same error diagnosis, error context determines different parameters for recovery actions
\end{enumerate}\vspace{-0.2cm}
Figure~\ref{fig:recovery-properties} shows the occurrence of these three situations in our developed recoveries. We most frequently use \textit{Immediate Actions}, to create short and responsive recovery actions to bring the task back to a known state. While less frequent, \textit{Shared Recoveries} aided in rapid development and \textit{Dynamic Recoveries} were crucial in creating a reactive system to deal with diverse faults.

The task executor and monitor use the following factors to determine what recovery to perform:\vspace{-0.2cm}
\begin{enumerate}
  \item \textit{Action/Task}: the location of the error in the task hierarchy
  \item \textit{Number of Aborts}: how many times the error has occurred without resolution
  \item \textit{Belief}: a subset of the robot's belief of the task, robot, or environment state
  \item \textit{Error Signal}: a specific error signal returned by a primitive action
  \item \textit{Immediate Action Result}: the result of actions taken for recovery execution
\end{enumerate}\vspace{-0.2cm}
Figure~\ref{fig:fault-diagnosis} illustrates that localizing the error within the task hierarchy was especially important in determining recoveries because the task and action reuse for different situations often required different recoveries. Overall, the use of a diverse factors shows that robust recovery requires a wide range of context.

Finally, Figure~\ref{fig:resumption-strategy} demonstrates that all resumption strategies described in Section~\ref{sec:monitor-resume} were necessary for designing a robust recovery system. The diversity in resumption strategies showcase the possibility of resuming from a task beyond simply re-attempting it entirely, and the use of resumption strategies other than \texttt{RESUME\_CONTINUE} show that resumption cannot always directly return to where the error occurred.

\subsubsection{Representative Example}

We describe here recoveries for the $pick$ action (Figure~\ref{fig:low-level-task-flow}) to provide concrete examples for the features mentioned above. In our system,  $pick$ and $arm$ can fail due to a common cause---errors in the MoveIt! Motion Planning Framework.  Therefore, the default recoveries for these actions, e.g. reinitializing a 3D obstacle map followed by a \texttt{RESUME\_CONTINUE}, are examples of \textit{Shared Recoveries}. However, depending on the context, the fault sometimes requires additional recovery steps. For instance, after the third consecutive failure of both actions in the $pick\_task$, a short upward arm move jogs the system out of its error. When this is not enough, the cause of failures can be a limitation in the arm's workspace, and so the robot repositions itself based on beliefs about the task and environment state. All faults that occur within the context of the $perceive\_pick$ task retry that task (\texttt{RESUME\_RETRY}) in order to account for scene changes resulting from recovery execution. We show a selection of these $pick$ recoveries in action, as well as other example recovery behaviors selected from our FetchIt! competition runs, in the video supplement to this paper\footnote{We also provide an HD version of the video here: \url{https://youtu.be/AcOdT10q_94}}.  At the competition, we recovered from all errors in the $pick$ task, thanks to the task monitor's utilities.
\vspace{-0.5cm}



\section{Discussion}
\label{sec:discussion}
\vspace{-0.3cm}

We conclude with a discussion of lessons learned using our RDD-inspired task execution and monitoring system at the FetchIt! Challenge.

\textit{Testing early and often.} The ability to repeatedly test the robot system in its target environment is a critical requirement for the robustness benefits of RDD--simulation alone will likely not lead to the same level of robustness. As such, RDD is not suited to hazardous or remote environments, such as space robotics. However, many target environments for robots are neither inaccessible nor catastrophically hazardous, and are therefore compatible with RDD. 

\textit{Stochasticity in behaviors.} As mentioned in Section~\ref{sec:monitor-diversity}, failure recovery requires a high degree of variety in recovery mechanisms. We have found that a degree of non-determinism at the robot's behavior level facilitates such recoveries. For instance, our sampling-based ranking approaches to object selection, grasp calculation, and place pose calculation provided the robot with successful alternatives when retrying actions after previous action attempts failed. 

\textit{Planning layer integration.} Recipe-based tasks can admit multiple recipes, which need to be selected or rescheduled at runtime based on factors such as time constraints or major execution errors. Predefined scripts, such as those created during RDD specification phases, cannot easily handle such situations.  The shortcoming can, however, be addressed by the higher level planning layer in the robot architecture. We found that the level of abstraction used in our hierarchical executive layer recipe scripts made the specification of our planning layer almost trivial. Additionally, the executor and monitor utilities we developed (e.g. beliefs) were a great help in the planning layer.

In conclusion, the RDD methodology of separating the nominal task specification from recovery specification provides numerous benefits, which our team validated at the FetchIt! Challenge.  Our use of RDD (1) allowed rapid development of the task and recoveries, (2) enabled independent testing and efficient re-use through abstraction for tasks and recoveries, (3) necessitated the development of system utilities that ultimately proved valuable in other aspects of system development, and most importantly, (4) afforded our system a level of robustness that would have been more difficult or time-consuming to achieve through other means.
\vspace{-0.5cm}

\section*{Acknowledgements}
\vspace{-0.3cm}
This work was supported by an Early Career Faculty grant from NASA's Space Technology Research Grants Program, NSF IIS 1564080, NSF GRFP DGE-1650044, and ONR N000141612835.

%
%
\vspace*{-0.3cm}
\bibliographystyle{spmpsci}
\bibliography{library_fetchit}

\end{document}